\documentclass[conference]{IEEEtran}
\IEEEoverridecommandlockouts

\usepackage{cite}
\usepackage{amsmath,amssymb,amsfonts}
\usepackage{booktabs}
\usepackage{array}
\usepackage{graphicx}
\usepackage{textcomp}
\usepackage{xcolor}
\usepackage{url}
\usepackage{pgfplots}
\usepackage{balance}
\usepackage[hidelinks]{hyperref}
\usepackage{microtype}
\pgfplotsset{compat=1.18}
\usepgfplotslibrary{groupplots}

\def\BibTeX{{\rm B\kern-.05em{\sc i\kern-.025em b}\kern-.08em
    T\kern-.1667em\lower.7ex\hbox{E}\kern-.125emX}}

\begin{document}

\title{Early Diagnosis of Wasted Computation in Multi-Agent LLM Systems via Failure-Aware Observability}

\author{
\IEEEauthorblockN{Xianyou Li}
\IEEEauthorblockA{New York University\\
New York, NY 10012, USA\\
xl4230@nyu.edu}
\and
\IEEEauthorblockN{Weiran Yan}
\IEEEauthorblockA{Independent Researcher\\
Santa Clara, CA 95051, USA\\
yanwr2016@gmail.com}
\and
\IEEEauthorblockN{Yichao Wu}
\IEEEauthorblockA{Northeastern University\\
Boston, MA 02115-5000, USA\\
wu.yicha@northeastern.edu}
\and
\IEEEauthorblockN{Penghao Liang}
\IEEEauthorblockA{Northeastern University\\
Boston, MA 02115-5000, USA\\
liang.p@northeastern.edu}
\and
\IEEEauthorblockN{Mengwei Yuan}
\IEEEauthorblockA{Independent Researcher\\
Milpitas, CA 95035-7246, USA\\
yuanmw1998@gmail.com}
\and
\IEEEauthorblockN{Jianan Liu}
\IEEEauthorblockA{Independent Researcher\\
Austin, TX 78613-5377, USA\\
jiananliu2408@gmail.com}
\and
\IEEEauthorblockN{Jing Yang}
\IEEEauthorblockA{Washington University in St. Louis\\
St. Louis, MO 63130, USA\\
jing.y@wustl.edu}
}

\maketitle

\begin{abstract}
Failure-aware observability diagnoses wasted computation in multi-agent LLM
systems before final-answer evaluation can explain what went wrong. We propose a
trace-based framework for a three-agent architecture---orchestrator, search
agent, and execution agent---that converts structured events into online signals
for loops, budget pressure, low information gain, and tool instability, then
adds offline semantic grounding metrics and selective LLM-as-judge evaluation.
On 165 GAIA validation traces under identical caps, 98 runs produce usable final
answers and 67 fail or stop without one. Among warned failed runs, 58.1\% of
tokens are spent after the first warning on average, indicating substantial
opportunity for intervention. A 10-task Level-2 pilot uses warnings to diversify
search or require evidence, reducing post-warning token fraction from 0.638 in
the baseline to 0.304. The results support a layered design: cheap online
signals help the orchestrator redirect or halt redundant behavior, while deeper
semantic checks identify whether completed answers are grounded enough to trust.
\end{abstract}


\section{Introduction}

Tool-using and multi-agent LLM systems interleave reasoning, retrieval, and
action \cite{yao2023react,schick2023toolformer,wu2023autogen}. This design can
increase task coverage, but it also creates long trajectories whose failures
are hard to interpret from final answers alone. Benchmarks such as GAIA
\cite{mialon2023gaia} measure final task success, yet a final score compresses
many different processes into one endpoint: redundant searches, unproductive
retrieval, execution errors, weak evidence, repeated coordination decisions, or
no final answer.

Recent work makes this diagnostic gap explicit. Faulty-agent analysis studies
how collaboration resilience depends on system structure
\cite{huang2024resilience}, while MAST provides a failure taxonomy, annotated
multi-agent traces, and an LLM-as-judge annotation pipeline
\cite{cemri2025why}. Related trajectory-centered work models agent paths with
large-and-small model collaboration \cite{du2025trajagentllmagentframeworktrajectory}.
Distributed-system tracing connects observed symptoms to internal execution
paths \cite{sigelman2010dapper},
but multi-agent LLM systems need a related form of observability: their costs
arise from model calls, retrieval, tools, and execution, while their failures
can be semantic rather than exceptional. Evaluation methods are also shifting.
LLM-as-judge methods provide semantic assessment but add model-call cost;
sentence embeddings offer cheaper semantic checks \cite{reimers2019sentencebert}.
Industry guidance likewise treats transcript records, graders, and tracked
metrics as core components of agent evaluation \cite{anthropic2026agentEvals}.

This paper treats wasted computation as an observability problem. We define
wasted computation as tokens, tool calls, computation attempts, and latency
spent on runs that do not produce a usable, grounded final answer. We use
cost-correlated proxies available during execution rather than direct hardware
energy or provider billing. Since agentic and test-time-scaling workflows can
increase inference energy demand \cite{Oviedo_2026}, the central claim is that
multi-agent LLM systems
require failure-aware observability in addition to final-answer evaluation.

We instantiate this claim in a three-agent question-answering system composed
of an orchestrator, a search agent, and an execution agent. Runs are stored as
structured event traces. Online signals identify repeated actions, low
information gain, missing evidence, and tool-failure streaks; offline analysis
derives reliability, grounding, semantic-support, and budget metrics from the
same records. Related systems improve agent fallback, memory, retrieval,
root-cause investigation, and iterative decision loops
\cite{11509181,11484130,11519347};
broader evaluation work shows that task labels can shift across platforms
\cite{Li_2021}. Our focus is orthogonal: trace-level signals that explain how
existing runs spend computation and where failure becomes visible.

The contributions of this paper are:
\begin{itemize}
    \item A failure taxonomy distinguishing execution failure,
    orchestration loops, evidence failure, no-final termination, and cost waste.
    \item A trace-based observability framework mapping failure modes to online
    metrics, offline semantic evaluations, and warning-triggered interventions.
    \item A level-stratified GAIA analysis showing how warnings, grounding
    signals, and budget proxies relate to failed computation.
\end{itemize}

\section{System and Trace Records}

\subsection{Agent Architecture}

The evaluated system has three roles: an orchestrator, a search agent, and an
execution agent. The orchestrator chooses among five structured actions:
seeking information, inspecting sources, using task materials, invoking bounded
execution, or producing a final response. The search agent returns source
summaries, while the execution agent processes task materials under controlled
runtime limits. All reported runs use deterministic, low-cost model settings.

\begin{figure*}[t]
\centering
\includegraphics[width=0.64\textwidth,trim=25 55 25 55,clip]{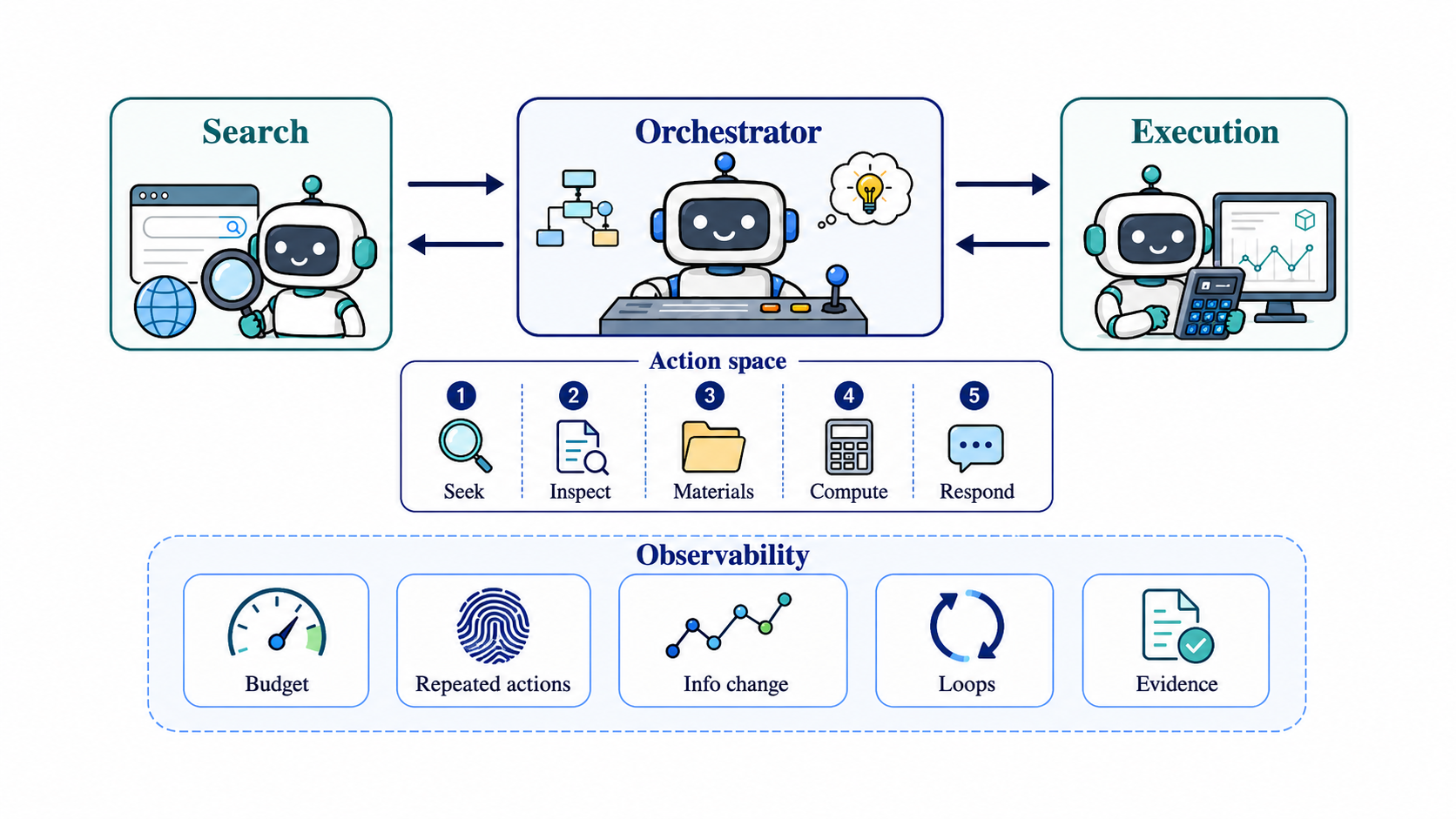}
\caption{Agent architecture and observability signals. The orchestrator
coordinates the search and execution agents through a five-action interface.
The observability layer summarizes the resulting trace along budget, repeated
action, information-change, loop, and evidence dimensions.}
\label{fig:agent-architecture}
\end{figure*}

Figure~\ref{fig:agent-architecture} separates the action interface from the
observability layer. The action space changes the run state; the observability
dimensions summarize what those actions produce. Seek and inspect actions feed
information-change and evidence signals, material and execution actions feed
budget and tool signals, and repeated normalized action signatures expose loops
across action types.

\subsection{Structured Trace Records}

Each run is represented as an ordered event record. Events contain a run
identifier, timestamp, event type, agent role, and structured payload, allowing
online warnings and offline metrics to be reconstructed from the trace itself.
The persistent storage backend is an implementation detail.

Table~\ref{tab:metric-map} summarizes the mapping from failure modes to
observable signals. It is read row-wise: each row names an operational symptom,
the signal layer, measurable trace features, and the corresponding triage
interpretation. These signals are not automatic stop rules; they indicate when
to trigger deeper evaluation, a behavioral nudge, or post-run analysis.

\begin{table*}[t]
\caption{Failure-aware observability map. Metrics are grouped by exposed
failure mode and signal layer.}
\label{tab:metric-map}
\centering
\footnotesize
\begin{tabular}{p{0.19\linewidth} p{0.15\linewidth} p{0.25\linewidth} p{0.31\linewidth}}
\toprule
\textbf{Failure mode} & \textbf{Layer} & \textbf{Observable signal} & \textbf{Operational interpretation} \\
\midrule
Tool instability & Search / execution & Tool error rate, retry summaries, latency & Calls are consuming budget without returning usable state. \\
Execution failure & Execution & Execution success rate, setup/timeout classes & External execution is not a reliable recovery path. \\
Repeated action loop & Orchestration & Repeated action keys, ABAB cycle labels, cache hits & The run is spending computation without changing strategy. \\
Low information gain & Orchestration & New URLs, extracted fact count, low-gain streaks & Search or page opening is no longer adding task-relevant state. \\
Evidence failure & Grounding & Evidence-present rate, citation consistency, answer-evidence similarity, sentence support & The final answer may be unsupported even if the run terminates. \\
Budget waste & System & Tokens, tool calls, budget pressure, post-warning remaining budget & Computation after an early warning is available for intervention or evaluation. \\
\bottomrule
\end{tabular}
\end{table*}

\section{Failure-Aware Metrics}

Let a run be $r$ and let $E_r$ denote its ordered event trace. We compute
metrics at two levels: online metrics emitted during the run and offline
metrics computed from the completed trace.

\subsection{Execution and Tool Reliability}

Tool error rate measures the fraction of tool results whose status is
\texttt{error}:
\begin{equation}
\mathrm{ToolErr}(r) = \frac{N_{\mathrm{err}}(r)}{N_{\mathrm{tool}}(r)}.
\end{equation}
Here $N_{\mathrm{tool}}$ is the number of tool-result events in the run and
$N_{\mathrm{err}}$ is the subset with error status. We separately measure
execution success and retry outcomes to distinguish unreliable tools from
ineffective recovery behavior.

\subsection{Orchestration and Loop Signals}

Orchestration signals characterize whether the agentic process is still
making progress. The system represents each action by a normalized signature
and uses repeated signatures or alternating repeated signatures as loop
proxies. It also records coarse information-gain signals, such as new source
references or newly extracted evidence. Repeated low-gain steps are treated as
stagnation indicators rather than as direct correctness judgments.

\subsection{Evidence and Grounding Signals}

Grounding metrics estimate whether a final answer is supported by the trace.
Always-on signals record evidence availability and citation consistency.
Deeper offline analysis estimates answer-evidence alignment with a lightweight
semantic encoder. The reported artifact uses a local LSA-style representation:
term-frequency vectors are projected into a lower-dimensional semantic space,
providing a cheap approximation of sentence relatedness without an additional
model call. The same interface can support Sentence-BERT-style embeddings
\cite{reimers2019sentencebert}. The sentence-level support proxy is:
\begin{equation}
\mathrm{Support}_{\tau}(r) =
\frac{1}{|S_a|}\sum_{s \in S_a}
\mathbb{1}\left[\max_{c \in C_e} \cos(f(s), f(c)) \ge \tau\right],
\label{eq:support}
\end{equation}
where $S_a$ are answer sentences, $C_e$ are evidence chunks, $f$ is the
embedding function, and $\tau=0.65$ in the reported analysis.

\subsection{Selective LLM-as-Judge Evaluation}

We use an LLM judge only as an offline grounding evaluation. For each evaluated
trace, the judge receives the original question, the final answer, and the
recorded evidence. A strict prompt asks whether the evidence supports the
answer and returns a support score in $[0,1]$. This evaluation measures grounding,
not benchmark correctness. The judge does not use prior knowledge to answer the
task and does not control the agent trajectory. We use deterministic decoding
and cache responses to avoid repeated calls.

\subsection{Wasted Computation Proxy}

For a configurable cost vector, run cost can be approximated as
\begin{equation}
C_r = \alpha T_r + \beta H_r + \gamma R_r + \delta X_r,
\label{eq:cost}
\end{equation}
where $r$ indexes a run, $T_r$ is total tokens, $H_r$ is tool calls, $R_r$ is
retries, $X_r$ is execution attempts, and $\alpha,\beta,\gamma,\delta$ are
deployment-specific unit-cost coefficients for these resources. These
coefficients convert heterogeneous resource counts into one comparable cost
scale; for example, a deployment could set them from provider token prices,
tool fees, measured latency, or energy accounting. A
wasted-computation ratio can then be defined as
\begin{equation}
\mathrm{WCR} =
\frac{\sum_r C_r \cdot \mathbb{1}[\mathrm{unusable}(r)]}
{\sum_r C_r}.
\label{eq:wcr}
\end{equation}
Here $\mathbb{1}[\mathrm{unusable}(r)]$ is an indicator that equals 1 when run
$r$ terminates without a usable final result and 0 otherwise. In the current
experiment, dollar-cost coefficients are left unset because provider pricing,
hardware allocation, caching, and tool billing vary across deployments. Setting
one artificial dollar vector would imply a billing model that the experiment
does not measure. We therefore report token and tool-call counts directly and
treat them as auditable computation-cost proxies rather than as monetary or
energy measurements.

\section{Experimental Setup}

We evaluate on the complete local GAIA 2023 validation snapshot: 165 tasks
split across benchmark-provided difficulty levels 1, 2, and 3 as 53/86/26
tasks. We treat the levels as increasing difficulty strata rather than as
separate datasets: level 1 contains the simplest tasks, level 2 is the largest
intermediate stratum, and level 3 is the smallest high-difficulty stratum in
the local snapshot. The final runs were configured with identical caps: 8
orchestrator steps, 8 orchestrator-paid tool calls, 30-second model-request
timeouts, deterministic decoding, and a 25,000-token soft ceiling. Trace
metadata records the task level, task
identifier, model settings, budget limits, timeout, batch name, and source-code
fingerprint; all reported level batches use the same recorded caps.

We define a usable final result operationally: the system must produce a final
response marked usable by the orchestrator. Failed or no-final outcomes include
explicit failed finals and traces without a final response. This definition
does not claim benchmark correctness; grounding quality and correctness are
evaluated separately.

We use two capped offline evaluations with \texttt{gpt-4.1-mini}, a model in
the GPT-4.1 family \cite{openai2025gpt41}. A 10-trace grounding evaluation
compares final answers with recorded evidence. A 30-trace correctness evaluation
samples usable finals across levels and compares them with the GAIA reference
answers. Both evaluations are cached and capped so that LLM-as-judge evaluation
remains a selective validation layer rather than an always-on metric.

\section{Results}

\begin{figure*}[t]
\centering
\begin{tikzpicture}
\begin{groupplot}[
    group style={group size=2 by 1, horizontal sep=1.15cm},
    width=0.425\textwidth,
    height=0.225\textwidth,
    ymin=0,
    grid=major,
    major grid style={draw=gray!18},
    axis line style={gray!55},
    tick label style={font=\scriptsize},
    label style={font=\small},
    title style={font=\small\bfseries},
    xtick={1,2,3},
    xticklabels={L1,L2,L3},
    xlabel={GAIA task level},
    enlarge x limits=0.25,
    nodes near coords,
    every node near coord/.append style={font=\scriptsize, color=black, yshift=2pt},
]
\nextgroupplot[
    title={Operational outcomes},
    ylabel={Runs},
    ybar=4pt,
    bar width=7pt,
    ymax=68,
    ytick={0,20,40,60},
]
\addplot+[fill=teal!55, draw=teal!80!black] coordinates {
    (1,31) (2,53) (3,14)
};
\addplot+[fill=orange!55, draw=orange!85!black] coordinates {
    (1,22) (2,33) (3,12)
};

\nextgroupplot[
    title={Mean token use},
    ylabel={Tokens},
    ymax=19000,
    ytick={0,4000,8000,12000,16000},
    scaled y ticks=false,
    yticklabel style={/pgf/number format/fixed, /pgf/number format/1000 sep={,}},
    every node near coord/.append style={font=\scriptsize, color=blue!70!black, yshift=3pt},
    nodes near coords style={/pgf/number format/fixed, /pgf/number format/precision=0, /pgf/number format/1000 sep={,}},
]
\addplot+[color=blue!70!black, mark=*, mark size=2.6pt,
          mark options={fill=white, line width=0.8pt}, very thick] coordinates {
    (1,8152.0) (2,9667.5) (3,16388.8)
};
\end{groupplot}

\end{tikzpicture}
\caption{Full level-stratified run under identical run caps. Usable finals are
operational outcomes, not correctness claims: 31/53, 53/86, and 14/26 runs are
usable at levels 1--3, with mean token use of 8,152, 9,668, and 16,389.
Teal bars denote usable finals; orange bars denote failed/no-final runs.}
\label{fig:partial-results}
\end{figure*}
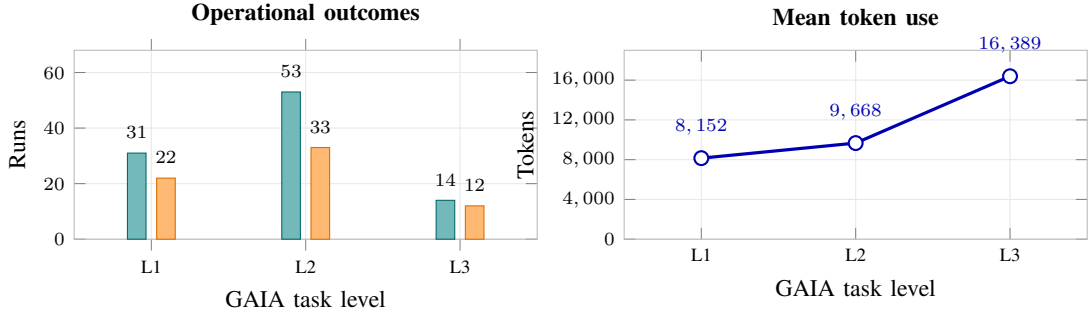

Figure~\ref{fig:partial-results} summarizes the complete validation-snapshot
run. On correctness, we report a selective offline evaluation over 30 runs that
produced usable finals (10 per level). Using \texttt{gpt-4.1-mini} as the
grader against GAIA validation reference answers, 12/30 evaluated usable finals
are judged correct (level-1: 6/10; level-2: 4/10; level-3: 2/10). We treat this
as a sanity check that separates ``usable final produced'' from benchmark
correctness; it is not a full accuracy estimate over all tasks or over failed
runs.

To quantify wasted computation after early warnings, we measure the fraction of
tokens consumed after the first triage label in failed runs. Across the 67
failed runs, 56 emitted at least one warning; among those warned failed runs,
the mean post-warning token fraction is 0.581 and the median is 0.611.

\subsection{Cross-Level Run Behavior}

Table~\ref{tab:level-summary} is a workload-level summary. \textit{Usable} and
\textit{Failed} are operational counts; the other columns are level means over
all runs. This reports the computation and signal prevalence seen online,
while later subsections analyze failure mechanisms and post-warning waste on
failed runs and grounding/correctness on final-bearing runs. \textit{Tools} is
mean orchestrator-paid tool calls; \textit{Support} is Eq.~\eqref{eq:support}
with $\tau=0.65$.

\begin{table*}[t]
\caption{Full level-stratified metrics. Support is the answer-sentence
fraction above the 0.65 evidence-similarity threshold.}
\label{tab:level-summary}
\centering
\footnotesize
\begin{tabular}{l r r r r r r r r r}
\toprule
\textbf{Level} & \textbf{Runs} & \textbf{Usable} & \textbf{Failed} &
\textbf{Triage} & \textbf{Evidence} & \textbf{Tool err.} &
\textbf{Tokens} & \textbf{Tools} & \textbf{Support} \\
\midrule
1 & 53 & 31 & 22 & 23 & 0.811 & 0.189 & 8,152.0 & 3.02 & 0.047 \\
2 & 86 & 53 & 33 & 44 & 0.977 & 0.103 & 9,667.5 & 3.71 & 0.201 \\
3 & 26 & 14 & 12 & 16 & 0.950 & 0.109 & 16,388.8 & 5.00 & 0.000 \\
\bottomrule
\end{tabular}
\end{table*}

Evidence availability is high in levels 2 and 3, but average support does not
increase monotonically because failed/no-final traces contribute little or no
answer support. Evidence presence, operational completion, and grounded
correctness should therefore be treated as related but separate outcomes.
Citation gaps remain visible: evidence is present without citations in 28.3\%,
36.0\%, and 46.2\% of levels 1--3.

\subsection{Failure Signals}

Trace labels separate unusable outcomes by mechanism. Level-1 failures include
insufficient evidence, loops, and step-limit termination; level-2 and level-3
failures are dominated by loops and step limits. Loop warnings appear in
16/31/9 runs across levels 1--3. Lower-frequency labels expose low information
gain, tool-failure streaks, evidence gaps, execution without usable output, and
blocked near-duplicate search.

\subsection{Wasted Computation After Warning}

For each run, we compare the token budget snapshot at the first triage label
with the final token count. Among warned failed runs, 54.5\%, 63.8\%, and
46.8\% of tokens in levels 1--3 are spent after the first warning on average,
showing that warnings often arrive early enough for budget-aware intervention.

\subsection{Intervention Pilot}

We evaluate a minimal warning-triggered policy on 10 level-2 tasks, chosen
because level 2 is the largest stratum and has many loop warnings. The policy
diversifies information seeking after loop warnings and requires opened
evidence after evidence-gap warnings. The pilot produces 8/10 usable finals;
the two failed runs end by step limit and repeated loop. Among warned failures,
the mean post-warning token fraction is 0.304 versus 0.638 in the full level-2
baseline. The comparison is directional because the pilot has one warned
failure, but it shows how warnings can trigger low-cost behavioral nudges.

\subsection{Trace Case Studies}

Aggregates show correlation; trace excerpts show how the signals arise.
Table~\ref{tab:trace-cases} compresses representative level-1--3 traces to the
decision-relevant sequence, exposed signal, and triage implication. The cases
generalize four patterns: repeated actions can waste budget despite successful
tools; usable finals can be weakly grounded; tool errors can recover when later
evidence is strong; and high-level tasks can keep spending tokens after
multiple warnings without reaching a final answer.

\begin{table*}[t]
\caption{Concrete trace case studies. Action order is preserved and actor roles
are shown explicitly; repeated environment outputs are compressed for
readability.}
\label{tab:trace-cases}
\centering
\scriptsize
\setlength{\tabcolsep}{3pt}
\begin{tabular}{>{\raggedright\arraybackslash}p{0.14\linewidth}
                >{\raggedright\arraybackslash}p{0.42\linewidth}
                >{\raggedright\arraybackslash}p{0.20\linewidth}
                >{\raggedright\arraybackslash}p{0.18\linewidth}}
\toprule
\textbf{Case} & \textbf{Role-labeled trajectory excerpt} & \textbf{Observed signal} & \textbf{Triage implication} \\
\midrule
Repeated loop
(\texttt{23dd907f}, L1) &
\textbf{Orch.:} seek Audre Lorde poem indentation
$\rightarrow$ \textbf{Search:} ok.
\textbf{Orch.:} repeat same seek
$\rightarrow$ \textbf{Search:} ok.
\textbf{Orch.:} inspect Poetry Foundation
$\rightarrow$ \textbf{Search:} ok; inspect same page
$\rightarrow$ ok; inspect Poeticous
$\rightarrow$ ok; inspect same page
$\rightarrow$ ok; inspect Poetry Foundation again
$\rightarrow$ ok. Final status: \texttt{repeated\_action\_loop}. &
\texttt{LOOPING\_WARNING};
\texttt{SEARCH\_REPEAT\_BLOCKED};
11,749 tokens; near-zero answer--evidence similarity. &
The run is active but strategically unchanged; the orchestrator should redirect
search or stop before more budget is spent. \\
\midrule
Weak grounding
(\texttt{dc22a632}, L1) &
\textbf{Orch.:} seek Ali Khan / New Mexican restaurant / James Beard
winners
$\rightarrow$ \textbf{Search:} ok; repeat and refine seek twice
$\rightarrow$ ok; inspect Civil Alchemy product page
$\rightarrow$ ok; inspect same page
$\rightarrow$ ok. Final status: \texttt{final\_ok}. &
Evidence present; no tool errors; 9,534 tokens; sentence support near zero;
answer--evidence similarity effectively zero. &
A clean final answer is not necessarily grounded; semantic evaluation or a
judge check should precede acceptance. \\
\midrule
Tool recovery
(\texttt{b2c257e0}, L2) &
\textbf{Orch.:} use task material three times
$\rightarrow$ \textbf{Search:} ok; seek U.S. ice-cream butterfat standards
$\rightarrow$ ok; inspect Wikipedia
$\rightarrow$ ok.
\textbf{Orch.:} invoke execution three times
$\rightarrow$ \textbf{Exec.:} errors. Final status: \texttt{final\_ok}. &
\texttt{TOOL\_FAILURE\_STREAK};
tool-error rate 0.333; support 0.682; answer--evidence similarity 0.633;
8,869 tokens. &
Tool failures should be interpreted with recovery context; strong later
grounding can justify continuation instead of immediate stop. \\
\midrule
Budget pressure
(\texttt{f5a8cc84}, L3) &
\textbf{Orch.:} use materials
$\rightarrow$ ok; seek NIH translation / Michaelis--Menten equation
$\rightarrow$ \textbf{Search:} ok; inspect article
$\rightarrow$ ok.
\textbf{Orch.:} invoke execution repeatedly
$\rightarrow$ \textbf{Exec.:} ok but no decisive final state. Final status:
\texttt{max\_steps\_exceeded}. &
\texttt{BUDGET\_PRESSURE\_LOW\_IG};
\texttt{LOW\_INFO\_GAIN\_STREAK};
\texttt{EXECUTION\_NO\_OUTPUT};
27,293 tokens; support 0.037. &
On harder tasks, warnings can arrive while substantial budget remains; repeated
low-gain computation should trigger redirection or budget gating. \\
\bottomrule
\end{tabular}
\end{table*}

\subsection{Metric-Layer Comparison}

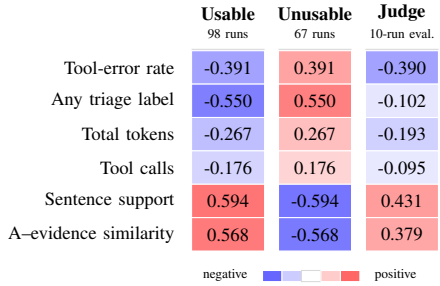
\begin{figure}[t]
\centering
\begin{tikzpicture}[x=0.92cm,y=0.92cm]
\scriptsize
\def\labelx{0.00}
\def\xone{2.80}
\def\xtwo{4.05}
\def\xthree{5.30}
\def\cellw{1.05}
\def\cellh{0.48}
\def\ystart{0.00}
\node[font=\scriptsize\bfseries,anchor=south] at (\xone+0.5*\cellw,0.34) {Usable};
\node[font=\tiny,anchor=south] at (\xone+0.5*\cellw,0.08) {98 runs};
\node[font=\scriptsize\bfseries,anchor=south] at (\xtwo+0.5*\cellw,0.34) {Unusable};
\node[font=\tiny,anchor=south] at (\xtwo+0.5*\cellw,0.08) {67 runs};
\node[font=\scriptsize\bfseries,anchor=south] at (\xthree+0.5*\cellw,0.34) {Judge};
\node[font=\tiny,anchor=south] at (\xthree+0.5*\cellw,0.08) {10-run eval.};

\newcommand{\heatrowval}[8]{%
  \node[anchor=east] at (\labelx+2.65,{\ystart-#1*\cellh-0.5*\cellh}) {#2};
  \draw[draw=white,fill=#4] (\xone,{\ystart-#1*\cellh-\cellh}) rectangle ++(\cellw,\cellh);
  \node at (\xone+0.5*\cellw,{\ystart-#1*\cellh-0.5*\cellh}) {#3};
  \draw[draw=white,fill=#6] (\xtwo,{\ystart-#1*\cellh-\cellh}) rectangle ++(\cellw,\cellh);
  \node at (\xtwo+0.5*\cellw,{\ystart-#1*\cellh-0.5*\cellh}) {#5};
  \draw[draw=white,fill=#8] (\xthree,{\ystart-#1*\cellh-\cellh}) rectangle ++(\cellw,\cellh);
  \node at (\xthree+0.5*\cellw,{\ystart-#1*\cellh-0.5*\cellh}) {#7};
}
\heatrowval{0}{Tool-error rate}{-0.391}{blue!36!white}{0.391}{red!36!white}{-0.390}{blue!36!white}
\heatrowval{1}{Any triage label}{-0.550}{blue!50!white}{0.550}{red!50!white}{-0.102}{blue!10!white}
\heatrowval{2}{Total tokens}{-0.267}{blue!25!white}{0.267}{red!25!white}{-0.193}{blue!18!white}
\heatrowval{3}{Tool calls}{-0.176}{blue!17!white}{0.176}{red!17!white}{-0.095}{blue!9!white}
\heatrowval{4}{Sentence support}{0.594}{red!54!white}{-0.594}{blue!54!white}{0.431}{red!39!white}
\heatrowval{5}{A--evidence similarity}{0.568}{red!52!white}{-0.568}{blue!52!white}{0.379}{red!35!white}

\draw[gray!30] (\xone,0.02) -- ++(\cellw,0);
\draw[gray!30] (\xtwo,0.02) -- ++(\cellw,0);
\draw[gray!30] (\xthree,0.02) -- ++(\cellw,0);
\node[anchor=east,font=\tiny] at (3.70,-3.22) {negative};
\draw[fill=blue!60!white,draw=white] (3.82,-3.32) rectangle ++(0.28,0.16);
\draw[fill=blue!20!white,draw=white] (4.10,-3.32) rectangle ++(0.28,0.16);
\draw[fill=white,draw=gray!30] (4.38,-3.32) rectangle ++(0.28,0.16);
\draw[fill=red!20!white,draw=white] (4.66,-3.32) rectangle ++(0.28,0.16);
\draw[fill=red!60!white,draw=white] (4.94,-3.32) rectangle ++(0.28,0.16);
\node[anchor=west,font=\tiny] at (5.34,-3.22) {positive};
\end{tikzpicture}
\caption{Pearson $r$ by metric layer. Operational-outcome columns use all 165
runs (98 usable, 67 unusable); judge-support correlations use the cached
10-trace grounding evaluation.}
\label{fig:metric-correlation}
\end{figure}

Figure~\ref{fig:metric-correlation} reports Pearson $r$ for all 165 runs
(98 usable, 67 unusable) and the 10-trace judge subset. Since usable and
unusable are complementary outcomes, their signs mirror. For unusable outcomes,
the largest cheap-signal association is any triage label ($r=0.550$), followed
by tool-error rate ($r=0.391$), total tokens ($r=0.267$), and tool calls
($r=0.176$). Semantic support and answer--evidence similarity associate with
usable outcomes ($r=0.594,0.568$) and with judge support ($r=0.431,0.379$).
These are descriptive correlations, not causal or accuracy estimates.

\section{Discussion, Limitations, and Conclusion}

The full run supports a layered observability design. Cheap trace signals expose
tool instability and stalled trajectories during execution; semantic metrics
separate evidence presence from answer support; selective LLM-as-judge
evaluation validates those proxies without becoming an always-on cost. Early
warnings are best treated as intervention opportunities rather than automatic
stop rules because some warned runs recover. A production system could change
retrieval strategy, require stronger evidence, reduce budget for low-probability
continuations, or route only ambiguous traces to a judge. This is why
final-answer accuracy alone is too coarse: no-final traces, weakly grounded
answers, and correct but redundant traces have different operational meanings.

\subsection{Limitations}

This study uses one agent configuration and one GAIA validation snapshot, so it
does not establish generality across models, prompts, or benchmarks. The
10-trace grounding evaluation and 30-trace correctness evaluation are
descriptive. Token and tool-call counts are cost proxies rather than direct
energy, latency, or billing measurements. Embedding similarity can miss
entailment and numerical correctness, and the intervention pilot is too small
to estimate accuracy-cost tradeoffs.

\subsection{Conclusion}

This paper introduced failure-aware observability for diagnosing wasted
computation in tool-using multi-agent LLM systems. On 165 GAIA validation
traces, unusable outcomes arise from evidence gaps, repeated loops, tool issues,
and step limits rather than one failure mode. Cheap metrics identify wasteful
trajectories early, semantic metrics deepen grounding analysis, and selective
LLM-as-judge evaluation validates the analysis without becoming always-on. The
next step is to measure controlled intervention policies for cost-adjusted task
success.

\end{document}